\documentclass{article}
\usepackage{iclr2027_conference,times}

\usepackage{amsmath,amsfonts,bm}

\def\eqref#1{equation~\ref{#1}}

\def\1{\bm{1}}

\DeclareMathAlphabet{\mathsfit}{\encodingdefault}{\sfdefault}{m}{sl}
\SetMathAlphabet{\mathsfit}{bold}{\encodingdefault}{\sfdefault}{bx}{n}

\usepackage{hyperref}
\usepackage{url}
\usepackage{booktabs}
\usepackage{array}
\usepackage{float}
\usepackage{multirow}
\usepackage{tabularx}
\usepackage{amssymb}
\usepackage[dvipsnames]{xcolor}
\usepackage{graphicx}
\usepackage{comment}
\usepackage{subcaption}
\usepackage{amssymb}
\usepackage{inconsolata}  
\usepackage[capitalize]{cleveref}
\usepackage{enumitem}
\usepackage{etoolbox}

\makeatletter
\patchcmd{\@maketitle}
  {\begin{tabular}[t]{l}\bf\rule{\z@}{24pt}\@author\end{tabular}}
  {\noindent\makebox[\textwidth][c]{%
    \begin{tabular}[t]{c}
      \bf\rule{\z@}{24pt}\@author
    \end{tabular}%
  }}
  {\typeout{ICLR author centering applied}}
  {\PackageError{iclr-centering}{Could not patch ICLR title}{}}
\makeatother

\hypersetup{
  colorlinks,
  linkcolor=red,
  citecolor=blue,
  urlcolor=magenta,
}

\definecolor{highlight}{RGB}{250,230,235}
\definecolor{bestgreen}{RGB}{226,239,218}

\makeatletter
\newcommand{\todo}[1]{\textcolor{red}{\textbf{TODO:}~#1}\@latex@warning{TODO: #1}}
\newcommand{\tocite}[1]{\textcolor{red}{#1~\textbf{[CITE]}}\@latex@warning{CITE: #1}}
\newcommand{\wip}[1]{{\color{gray}#1}\@latex@warning{WIP segment}}

\definecolor{daniel}{rgb}{0.8,0.4,0.0}   
\definecolor{tristan}{rgb}{0.7, 0.3, 0.7}  

\newcommand{\coloredcomment}[2]{{\color{#1}#2}\@latex@warning{#2}}

\makeatother

\newcommand{\diff}{\mathop{}\!\mathrm{d}}
\newcommand{\PertVec}{\boldsymbol{\gamma}}

\title{Attribution-based diagnostics for\\single-cell foundation models}

\author{%
  \textbf{Farnoush Rezaei Jafari}$^{1,2,\ast}$ \quad
  \textbf{Tristan Lazard}$^{3}$ \quad
  \textbf{Kenza Bouzid}$^{3}$ \\[0.4em]
  \textbf{Shruthi Bannur}$^{3}$ \quad
  \textbf{Jeremiah Wander}$^{3}$ \quad
  \textbf{Ashley Mae Conard}$^{3}$ \\[0.4em]
  \textbf{Stephanie Hyland}$^{3}$ \quad
  \textbf{Lorin Crawford}$^{3}$ \quad
  \textbf{Daniel Coelho de Castro}$^{3,\ddagger}$ \\[0.8em]
  $^{1}$Technical University of Berlin \\
  $^{2}$BIFOLD -- Berlin Institute for the Foundations
  of Learning and Data, Berlin, Germany \\
  $^{3}$Microsoft Research
}

\iclrfinalcopy 

\begin{document}
\maketitle

\vspace{-0.35in}
\begin{center}
\footnotesize
$^{\ast}$Work done during an internship at Microsoft Research.\\
$^{\ddagger}$Corresponding author:
\texttt{dacoelh@microsoft.com}
\end{center}
\vspace{0.20in}

\begin{abstract}
Single-cell RNA foundation models (scFMs) are increasingly used for \textit{in-silico} biological discovery, yet predictive performance alone does not reveal whether their decisions rely on biologically meaningful signals or model-specific shortcuts.
We systematically investigate post-hoc attribution of downstream predictions across two tasks: cell-type annotation using pretrained Cell2Sentence and scGPT models, and perturbation prediction using scGPT+STATE in a controlled setting.
For cell annotation, we use attribution to assess whether predictions rely on established marker genes and how their relevance varies with model architecture, expression rank, and the surrounding cellular context.
 For perturbation prediction, we compare attribution patterns with known regulatory ground-truth in a synthetic setting to assess what they reveal about the model's learned computations.
Across these tasks, attribution reveals biological feature use, contextual interactions, contributions from different input pathways, learning progression, associations with prediction quality, and potential pitfalls that are not fully captured by standard performance metrics.
Experiments with complete and limited regulatory knowledge further demonstrate that conclusions drawn from attribution depend strongly on the choice of biological reference, baseline, and null model.
Together, these findings highlight the value of attribution as a flexible post-hoc diagnostic for understanding model behavior, identifying potential issues, and informing the development of scFMs.
\end{abstract}

\section{Introduction}

The development of single-cell foundation models (scFMs) is transforming biological research, with the promise of accelerating discovery through \textit{in silico} experimentation~\citep{jiang2026landscape}. However, extrapolation remains challenging. Generalizing beyond the training distribution requires models to capture biological mechanisms that remain stable across changing conditions rather than relying on spurious correlations~\citep{tejada-lapuerta2025causal}.
This motivates methods for assessing whether the internal computations underlying scFM predictions are biologically plausible, both to better understand model failures and to evaluate whether successful predictions arise for biologically meaningful reasons~\citep{tejada-lapuerta2025causal,dimitrov2026singlecells}.

Interpretability provides a natural framework for such analysis. Beyond supporting model debugging and trust~\citep{rauker2023transparent}, inspecting learned representations and computations can provide evidence for learned internal models of the underlying domain~\citep{li2025worldmodel} and may itself serve as a tool for scientific discovery~\citep{freiesleben2024scientific}. Existing interpretability studies of scFMs have primarily analyzed intermediate representation spaces, identifying biological information encoded in cell or gene embeddings (see \cref{sec:related_work}). However, the presence of biological information in a representation does not reveal which input features support that information or how it is used by downstream predictors built on top of the representation. A more direct analysis therefore requires linking downstream predictions back to biologically meaningful input features.

\begin{figure}[t]
    \centering
    \includegraphics[width=\textwidth]{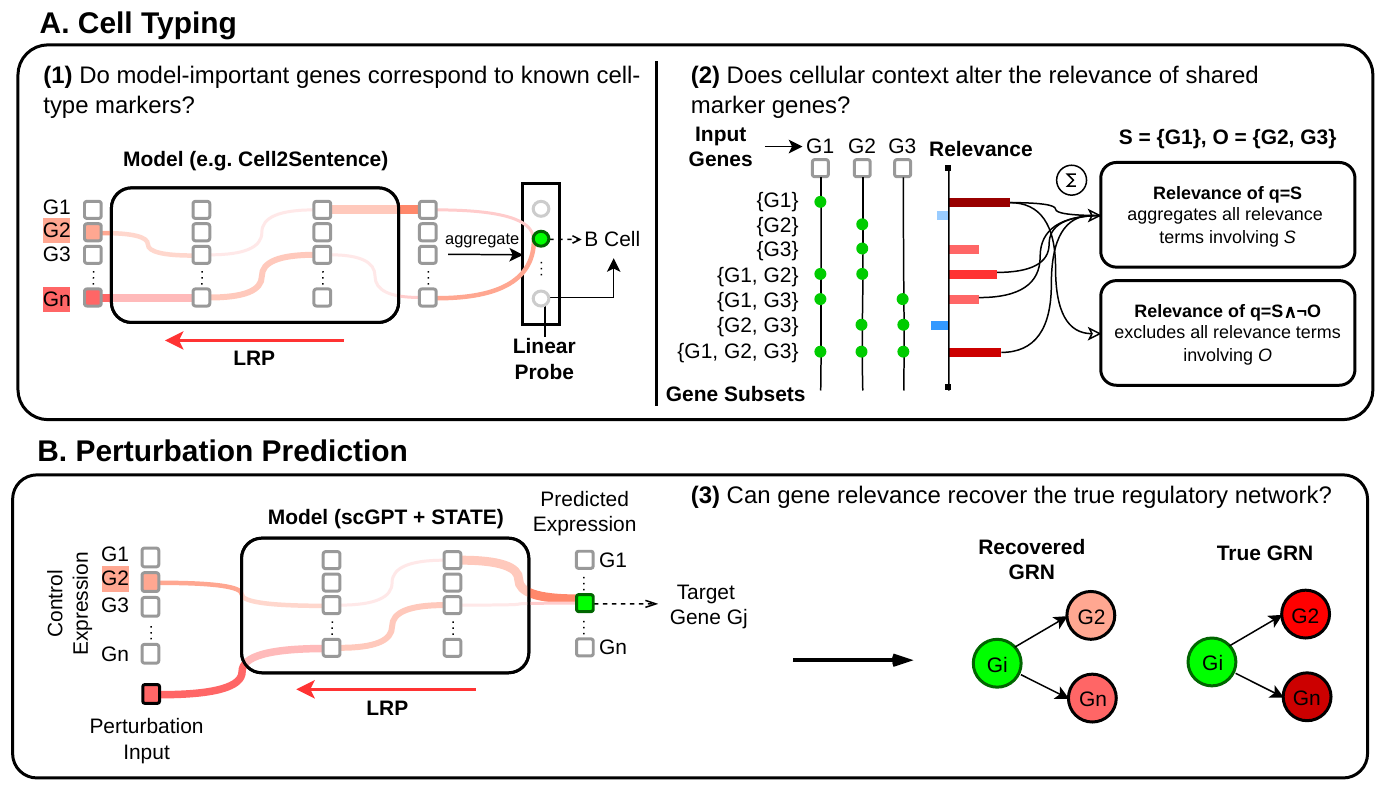}
    \caption{
    Overview of our framework for explaining downstream predictions of single-cell foundation models.
    \textbf{A.1.} Layer-wise Relevance Propagation (LRP) decomposes a cell-type probe logit into per-gene relevance scores, which are compared with known markers of the predicted cell type.
    \textbf{A.2.} Symbolic XAI decomposes the probe logit into relevance scores for gene subsets, which can be combined into logical queries to quantify how the relevance of marker genes changes in different contextual settings.
    \textbf{B.3.} For each perturbation of gene \(G_i\) and output gene \(G_j\), LRP attributes the predicted expression of \(G_j\) to the model inputs, quantifying the contribution of \(G_i\). Repeating this across perturbations yields a gene--gene relevance matrix that can be interpreted as an attribution-based estimate of regulatory influence and compared with the ground-truth gene regulatory network (GRN).
    Red paths in A and B indicate relevance propagation through the models, with line width and color intensity encoding relevance magnitude.
    }
    \label{fig:overview}
\end{figure}

In this work, we use attribution-based explainability to assess whether downstream predictors built on scFM representations rely on biologically meaningful signals consistent with established biological knowledge, as illustrated in \cref{fig:overview}.
We study two common applications of increasing complexity. For cell type annotation, we assess whether predictions rely on established cell-type marker genes beyond confounding factors such as expression rank, and use symbolic explanations to examine how the relevance of marker genes shared by closely related cell subtypes changes in different contextual settings.
For perturbation prediction, we assess how closely attribution patterns align with the underlying gene regulatory network, providing a diagnostic of what the model has learned. Because reliable regulatory ground-truth is generally unavailable in real perturbation datasets, we construct a synthetic setting with a fully specified causal regulatory system, enabling direct comparison between model explanations and the mechanisms used to generate the data.
Our contributions are threefold:
\begin{enumerate}[label=(\arabic*)]
    \item We provide the first systematic study of post-hoc scFM explainability that focuses on dissecting downstream task predictions rather than only analyzing information encoded in intermediate representations or attention patterns.

    \item We develop attribution-based analyses that connect downstream predictions to established biological knowledge at both the level of individual genes and higher-order contextual interactions  (\cref{sec:cell_typing}).
    
    \item We demonstrate that attribution-based explainability can serve as a practical tool for analyzing gene regulatory structure learned by scFMs, enabling direct evaluation of whether model attributions recover known regulatory relationships in a controlled perturbation setting (\cref{sec:perturbation_prediction}).
\end{enumerate}

\section{Related Work}
\label{sec:related_work}

\paragraph{Post-hoc interpretability of biological foundation models.}
Interpretability has been increasingly used to investigate the biological information learned by foundation models. In protein language models, supervised probing of learned embeddings has been used to evaluate the decodability of biological attributes at both residue and protein levels~\citep{li2024reuse,whitfield2026highresolution}, while intervention-based analyses have revealed three-dimensional contacts even in models trained only on protein sequences~\citep{nainani2025mechanistic,candido2026language}. Unsupervised sparse autoencoders have further been used to decompose protein-model representations into more interpretable features~\citep{rives2021esm1b,adams2025mechanistic,simon2025interplm,gujral2025sparse,nainani2025mechanistic,candido2026language}, revealing features associated with structural, physicochemical, and functional properties. Beyond proteins, interpretability analyses of DNA foundation-model embeddings have also uncovered phylogenetic structure~\citep{pearce2025tree}.

\paragraph{Post-hoc interpretability of single-cell foundation models.}
Within transcriptomics, recent work has largely examined the representations learned by scFMs. Sparse autoencoders have uncovered cell- and gene-level features associated with known pathways and marker genes~\citep{claye2025discovering,pedrocchi2025sparse}, while zero-shot analyses have studied how embeddings organize biological properties such as cell identity relative to technical factors~\citep{kedzierska2025zeroshot}. Attribution methods have further estimated the influence of individual genes on learned representations~\citep{zhao2026how,gold2026signature} or interpretable features derived from them~\citep{claye2025discovering}. Attention weights have also been used to rank genes for annotated cell types~\citep{yang2022scbert} and to investigate recovery of regulatory interactions~\citep{kendiukhov2026systematic}. However, attention is output-agnostic and captures only one step of the computation, so it need not reflect what drives a prediction ~\citep{jain2019attention,wiegreffe-pinter-2019-attention}.

These studies show that scFM representations encode biologically meaningful information, but largely focus on intermediate representations. In contrast, we explain \emph{downstream task predictions} by propagating a chosen output through the full network with LRP~\citep{bach-plos15}, assigning each input feature a signed relevance for that prediction. While \citet{keyl2023single} use LRP on a shallow, task-specific network trained to impute masked gene expression to construct cell-specific regulatory networks, we use it to audit downstream predictions from pretrained scFMs. This allows us to test whether biological signals are not only represented, but actually used for prediction. We evaluate these explanations against explicit biological hypotheses, including known cell-type markers and the ground-truth gene regulatory network in a controlled perturbation setting.

\section{Cell type classification}
\label{sec:cell_typing}

Cell typing is a fundamental task in single-cell analysis, where each cell is assigned to a biologically meaningful cell type based on its gene-expression profile. 
Although single-cell foundation models can achieve strong cell-type classification performance, predictive accuracy alone does not reveal what biological information the models use to distinguish cell populations.
We therefore use cell typing as our first interpretability setting, asking whether model predictions are driven by biologically meaningful signals.

\subsection{Experimental setup}
\paragraph{Datasets.}
We curate two benchmarks from the CZ CELLxGENE Census \citep{czi2025cellxgene} with well-characterized molecular signatures, comprising peripheral blood cells and adult human brain nuclei. Each sample is represented by its gene expression profile.
The blood dataset has a two-level label hierarchy, with four coarse cell types (T cells, B cells, natural killer cells, and monocytes),
further divided into 22 subtypes. The brain dataset contains seven broad cell types, including excitatory and inhibitory neurons, astrocytes, oligodendrocytes, oligodendrocyte precursor cells, microglia/macrophages, and vascular/perivascular cells.
Both datasets contain $\sim$61K samples, and are split into train, validation, and test sets using a 70/15/15 ratio.
Further dataset details are provided in \cref{app:datasets}.

\paragraph{Models and context settings.}
\label{sec:models}
We evaluate two families of pretrained single-cell foundation models throughout this study, Cell2Sentence \citep[C2S;][]{levine2024cell2sentence,rizvi2026scaling} and scGPT~\citep{cui2024scgpt}.
C2S models are adapted from pretrained autoregressive language models (here, Pythia-410M, Pythia-1B, Gemma2-2B). They represent cells as sequences of top-$K$ genes sorted by expression value, tokenized with the original text vocabulary.
We use $K=200$ for C2S (Pythia-410M) and both $K=200$ and $K=1000$ for the larger C2S-Scale models, Pythia-1B and Gemma2-2B, allowing us to examine the effects of model scale and input context.
scGPT models are encoder-only Transformers that represent gene names and their expression values as combined token embeddings and were trained with a maximum context length of $K=1200$. We consider scGPT models pretrained on blood, brain, and whole-human cells.

\paragraph{Cell-type classification with linear probes.}
To obtain task-aligned predictions without modifying the pretrained scFMs, we freeze each model and train linear probes on final-layer representations to predict blood cell types and subtypes, and brain cell types.
This is necessary even for C2S models that can generate cell-type labels directly, since their generated labels may not match the annotation granularity of our datasets.
For C2S, we compare mean pooling over non-padding tokens with the final-token representation, and for scGPT mean pooling with the global cell representation (\texttt{[CLS]}), selecting the better-performing option per model configuration and task based on validation performance.
The probes achieve high accuracy overall, with near-perfect performance on coarse cell types.
Hyperparameters and pooling/layer sweep results are detailed in~\cref{app:probe-details}.

\subsection{Marker recovery beyond expression value}
\label{sec:marker_recovery}

To assess whether linear-probe predictions are supported by biologically meaningful evidence, we examine whether the genes most relevant to each prediction align with established cell-type markers rather than merely reflecting expression magnitude. 
We use Layer-wise Relevance Propagation \citep[LRP;][]{bach-plos15} to assign each input gene a signed relevance score indicating how strongly it supports or opposes the probe prediction. LRP obtains these scores by propagating the model output backward through the computational graph.%
\footnote{For C2S models, the relevance of a gene is obtained by summing the scores over its name's sub-tokens.}
Please refer to \cref{app:lrp} for details.

We evaluate marker recovery using average precision (AP), which measures how strongly marker genes are concentrated near the top of a gene ranking. For each cell, genes in the marker set \(\mathcal{M}\) are treated as positives and all remaining genes as negatives. Genes are ranked by their relevance scores, and AP is computed from the precision--recall curve as \(\mathrm{AP}=\sum_i (R_i-R_{i-1})P_i\), where \(P_i\) and \(R_i\) denote precision and recall at the \(i\)-th score threshold. Tied scores are evaluated jointly, making AP invariant to their input ordering. We compute AP per cell using signed gene-level relevance scores and, as a model-free baseline, gene-expression values. Scores are then macro-averaged across biological classes, with correct and incorrect predictions reported separately.

\begin{figure}[tb]
    \centering
    \includegraphics[width=0.9\linewidth, trim={0 10pt 0 8pt}, clip]{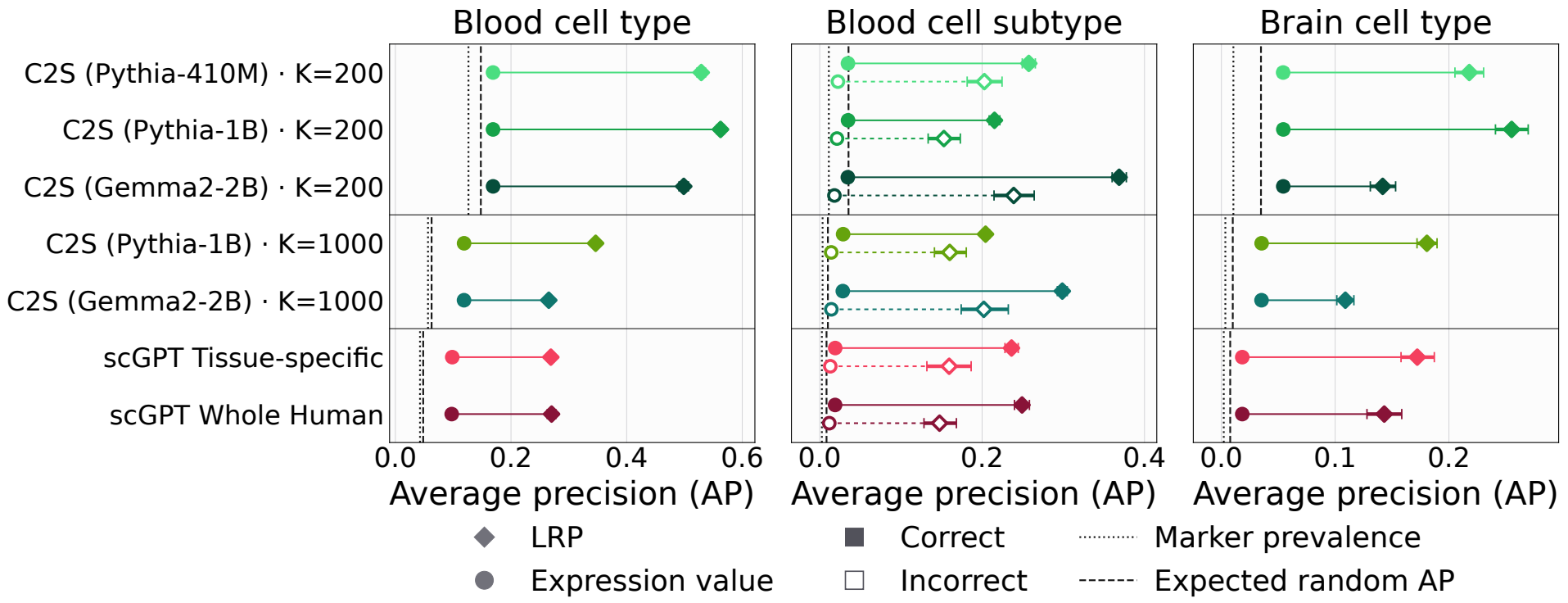}
    \caption{Mean marker-recovery average precision (AP) across tasks and models. Class-stratified bootstrapping 95\% CIs are also shown, though many are too narrow to visualise at this scale.
    Relevance scores recover marker genes better than expression values and random ranking.
    For blood cell subtypes, AP is higher for correct than incorrect predictions, indicating that prediction errors are associated with weaker marker alignment.
    Incorrect predictions are reported only for blood subtypes, since each model makes at most five errors on the broader cell-type tasks.
    Marker prevalence and expected AP under random finite rankings are shown for reference. 
    }
    \label{fig:marker_recovery}
\end{figure}

\Cref{fig:marker_recovery} shows that relevance scores recover marker genes above chance across models and tasks. For blood cell subtyping, \(\mathrm{AP}_{\mathrm{LRP}}\) is consistently higher for correct than incorrect predictions, suggesting that errors are associated with weaker alignment between relevance  scores and known marker genes.
The expression-value baseline varies little across context settings, suggesting that most marker genes are already concentrated among the top \(K=200\) highly expressed genes.
%
Relevance scores for both C2S and scGPT models dramatically outperformed the expression-value baseline.
C2S (Gemma2-2B) configurations show the strongest marker recovery for blood cell subtyping, but slightly weaker recovery than Pythia configurations for blood and brain coarse cell types, implying a nuanced effect of model scale and architecture.

\paragraph{Position bias in C2S.}
C2S represents each cell as a sequence of gene identifiers sorted by expression value, so token position reflects expression magnitude. 
Because the language model backbone also uses position through positional encodings and causal attention, biological relevance and sequence position are confounded, meaning a gene may be relevant because it is informative or because it appears early. 
To examine this effect, we use $K=200$ genes and divide each input sequence into 10 equally-sized bins.
For each cell, we normalize the gene-level relevance scores by the maximum absolute relevance score within that cell, sum the normalized scores within each bin, and then aggregate these bin-level scores across cells.

\begin{figure}[tb]
    \centering
    \includegraphics[width=0.95\linewidth, trim={0 10pt 0 8pt}, clip]{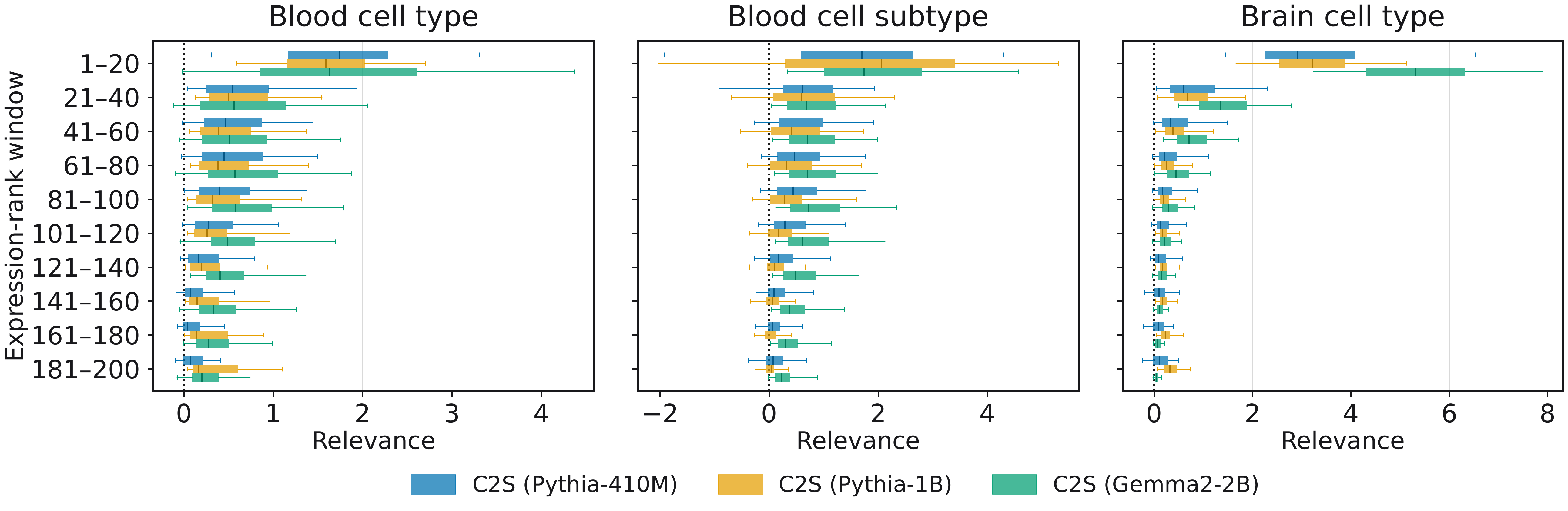}
    \caption{
    Distribution of relevance scores across gene-rank bins, ordered by decreasing expression. Box plots show the distribution across test cells. Relevance is concentrated among early, highly expressed genes and decreases with rank, most strongly for C2S (Gemma2-2B), although this pattern alone does not establish positional shortcut behavior.
    }
    \label{fig:c2s_gene_position_relevance}
\end{figure}

\Cref{fig:c2s_gene_position_relevance} shows that relevance is strongly concentrated in the earliest positions and declines steadily with rank across all models, most prominently in C2S (Gemma2-2B). This pattern alone does not establish positional shortcut behaviour, since early positions correspond to highly expressed genes that may themselves be biologically informative.

\subsection{Contextual relevance of shared marker genes via symbolic queries}
\label{sec:contextual}

\begin{figure}[tbh]
    \centering
    \includegraphics[width=0.85\linewidth]{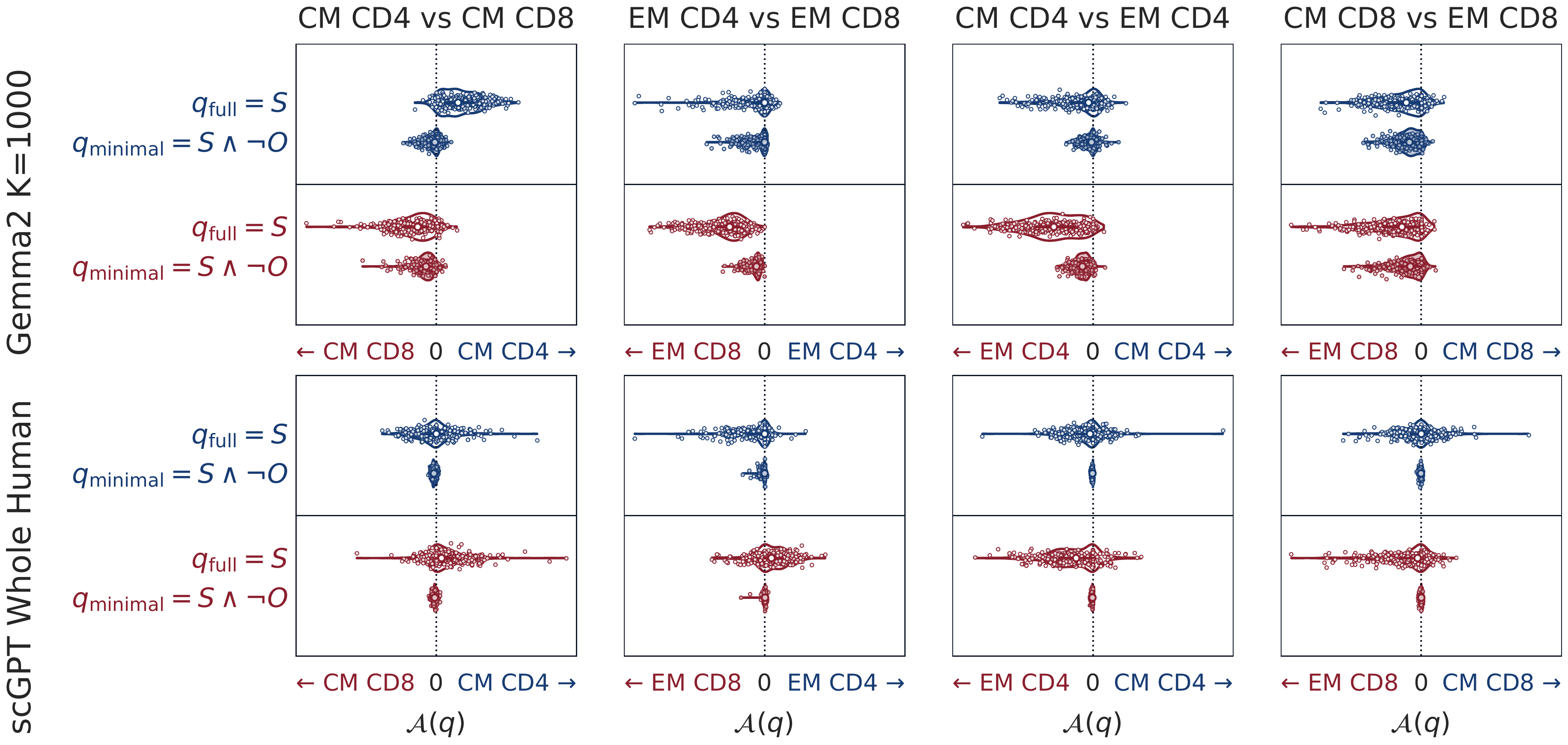}
    \caption{Context-sensitivity of C2S (Gemma2-2B) with $K=1000$ and scGPT Whole Human using symbolic queries.
    The relevance of $q_{\mathrm{minimal}}$
    is consistently concentrated near zero, whereas the full-context query
    $q_{\mathrm{full}}$ shows broader and often shifted relevance.
    This indicates that shared markers derive much of their relevance from
    interactions with the surrounding cellular context rather than from
    contributions confined to the shared-marker set.
    }
    \label{fig:context-sensitivity}
\end{figure}

Marker recovery shows whether predictions rely on biologically relevant genes, but not whether this relevance depends on the surrounding cellular context. This is important for closely related cell subtypes that share marker genes, whose discriminative value may emerge only through interactions with other genes. We use Symbolic XAI~\citep{schnake2025symbolic}, which quantifies the relevance of higher-order, human-interpretable logical queries, composed of input features and  operators like $\neg$, $\lor$, and $\land$, to assess whether shared markers receive context-dependent relevance. Refer to~\cref{app:symbolic_xai_details} for further details.

Let $\mathcal{S}$ be the set of shared markers and $\mathcal{O}$ the remaining tokens in the input.
We define two queries over the shared markers: \(q_{\mathrm{full}}=\mathcal{S}\), which measures the relevance of the shared markers in the full cellular context, and \(q_{\mathrm{minimal}}=\mathcal{S}\land\neg\mathcal{O}\), which measures their relevance excluding interactions with genes outside the shared-marker set.
The difference between the relevance assigned to these queries quantifies the extent to which the model is context-sensitive.

We study four pairs of closely related blood cell subtypes, each sharing two marker genes (\cref{tab:shared-marker-pairs}), using C2S (Gemma2-2B, \(K=1{,}000\)) and scGPT (Whole Human). We restrict the analysis to correctly classified cells and explain the logit difference between each subtype pair~\citep{contrastivelrp23}, such that positive relevance supports one subtype and negative relevance the other.

\Cref{fig:context-sensitivity} shows a clear separation between the two queries. The relevance of $q_{\mathrm{minimal}}$ remains near zero across subtype pairs, whereas $q_{\mathrm{full}}$ is substantially broader and often shifted. This suggests that shared marker genes provide little subtype-specific evidence on their own and become informative mainly through interactions with the surrounding cellular context. 
For C2S, some shifts align with how strongly the shared markers are represented in the input. In EM CD4 versus EM CD8, \textit{CCL5} and \textit{GZMA} occur much more often in EM CD8 cell sentences, consistent with the shift of $q_{\mathrm{full}}$ toward EM CD8. Likewise, in CM CD4 versus EM CD4, \textit{IL7R} is more abundant and higher ranked in EM CD4, matching the stronger full-context relevance toward EM CD4.

\section{Perturbation prediction}
\label{sec:perturbation_prediction}

Having first examined cell typing as a relatively simple setting for assessing whether predictions rely on biologically meaningful signals, we extend our analysis to genetic perturbation prediction, a more complex task with important applications in areas such as drug discovery. This progression allows us to move from cell-identity signals to the regulatory dependencies that govern cellular responses to intervention. Recent benchmarks suggest that scFMs do not consistently outperform simpler baselines on perturbation prediction \citep{ahlmann-eltzeDeeplearningbasedGenePerturbation2025,kernfeldComparisonComputationalMethods2025, bendidiBenchmarkingTranscriptomicsFoundation2024}, motivating a closer examination of what biological structure these models capture beyond predictive performance. We therefore use attribution-based analysis to assess whether the learned dependencies reflect the underlying gene regulatory network.

\subsection{Experimental setup}

\paragraph{Synthetic dataset.}
Ground-truth regulatory relationships are often unavailable in real single-cell perturbation datasets, making direct evaluation of whether model predictions reflect the underlying gene regulatory network (GRN) difficult. We therefore use a synthetic setting with fully known regulatory structure. Following \citet{qi2026mechanisms,lorch2026latent}, cells are generated from context-specific linear stochastic differential equations followed by a noisy emission process, yielding two cellular contexts over 128 genes.
We generate 8M unperturbed control cells for pretraining and, for fine-tuning, an additional 5M controls and 1M perturbed cells across both contexts. The perturbed set contains 8,192 independently simulated cells for each of 128 randomly signed single-gene perturbations, covering all genes. Controls and perturbed cells are sampled independently from their respective stationary populations and thus have no one-to-one correspondence.
Refer to \cref{app:synthetic_data_details} for full details.

\paragraph{Model.}
For perturbation analysis, we use scGPT as the backbone in a two-stage training procedure. We first pretrain scGPT from scratch on control cells using masked gene-expression reconstruction~\citep{cui2024scgpt}. We then feed its cell embeddings to the State Transition architecture of STATE~\citep{adduri2025state} and fine-tune the joint model on perturbation data. STATE processes sets of control cells and mixes their intermediate representations through self-attention to predict distributions of perturbed cell states. Details are provided in \cref{app:scgpt_training_details}. To distinguish learned behavior from architectural inductive biases, we also evaluate two random-weight baselines with five replicates each, fully random initialization and pretrained scGPT with an untrained STATE head.
The model preserves perturbation identity (centroid accuracy (\(0.962\))) and captures dominant response directions (Pearson \(\Delta_{20}=0.704\)), despite worse all-gene RMSE than the perturbed-mean baseline (\(0.855\) vs.\ \(0.606\)). Full results and metric definitions are provided in \cref{app:perturb_metrics,app:extended_results}.

\subsection{Recovering Regulatory Relationships with LRP}
\label{sec:grn_recovery}

We use LRP to test whether scGPT+STATE has learned the direct and indirect gene--gene relationships encoded in the synthetic GRN. For each perturbed gene, the ground-truth GRN specifies which other genes it directly regulates. We therefore ask whether highly relevant genes coincide with these known direct targets. 

All quantities below are evaluated separately for each cellular context, and we omit the context index \(c\) for readability.
Let \(G\) denote the number of genes. The simulator provides a regulatory matrix $\*A \in\mathbb{R}^{G\times G}$, where off-diagonal entry $A_{j,i}$ represents direct regulation $i \rightarrow j$. 
A perturbation $q$ targets gene $i_q$ and is represented by a signed, sparse intervention vector $\PertVec_q \in \mathbb{R}^G$, non-zero only at index $i_q$.
We consider two ground-truth perturbation effect vectors:
\begin{equation}
    \*d_q = \*A\PertVec_q, \qquad \*t_q = -\*A^{-1}\PertVec_q, \qquad \*d_q, \*t_q \in \mathbb{R}^G
\end{equation}
$\*d_q$ encodes the direct effects of perturbation $q$ on genes directly regulated by $i_q$.
Under the simulator's linear dynamics, the total stationary effect $\*t_q$ gives the expected change in latent cell state \citep{qi2026mechanisms} and incorporates direct and indirect effects.
For each perturbation \(q\), a signature \(m\) yields a gene-level score vector \(\*s_{m,q}\in\mathbb{R}^G\), where entry \(j\) scores output gene \(j\) after averaging over cells and STATE batches. We quantify the \emph{regulatory alignment} of each signature by comparing \(\*s_{m,q}\) with the ground-truth direct-effect vector \(\*d_q\) and total stationary-effect vector \(\*t_q\).

\paragraph{LRP signatures.}
We consider two LRP signatures. \textit{Perturbation relevance} measures how much the STATE perturbation input contributes to each predicted expression value. \textit{Same-cell gene LRP} propagates relevance backward through scGPT to the input genes of the corresponding control cell.
The signatures are evaluated to be faithful to the model and robust to the choice of LRP rules (\cref{app:lrp}).
Same-cell gene LRP initially yields a $G\times G$ array per perturbation, indexed by input and output genes. To reduce it to a signature $\*s_{m, q}$, we keep only the input entries corresponding to the perturbed gene $i_q$. These signatures measure contributions of that gene expression in the control state, rather than the effect of the perturbation. 

\paragraph{Baselines.}
As baselines, \textit{Empirical DE} is the differential expression between observed perturbed and control cells and serves as an oracle upper-bound for the regulatory signal present in the data. \textit{Predicted DE} is the corresponding differential expression computed from the model predictions and serves as a black-box baseline for the regulatory information captured in the model output. We also evaluate two black-box counterfactuals: ablating the perturbation input and masking gene $i_q$ (\cref{app:extended_regulatory_alignment}).

\paragraph{Random control.}
For each context, we construct a null distribution from 999 randomly rewired versions of its GRN that preserve each gene's in- and out-degree. We keep the signatures fixed and recompute the same recovery metrics against each randomized network, macro-averaged over perturbations. One-sided empirical \(p\)-values quantify how often the randomized GRNs achieve a score at least as high as the ground-truth GRN. The exact definition is provided in~\cref{app:random_controls}.

\subsection{Regulatory alignment beyond differential expression}
\label{sec:grn_alignment}

\begin{figure}[t]
    \centering
    \includegraphics[width=\linewidth, trim={0 10pt 0 8pt}, clip]{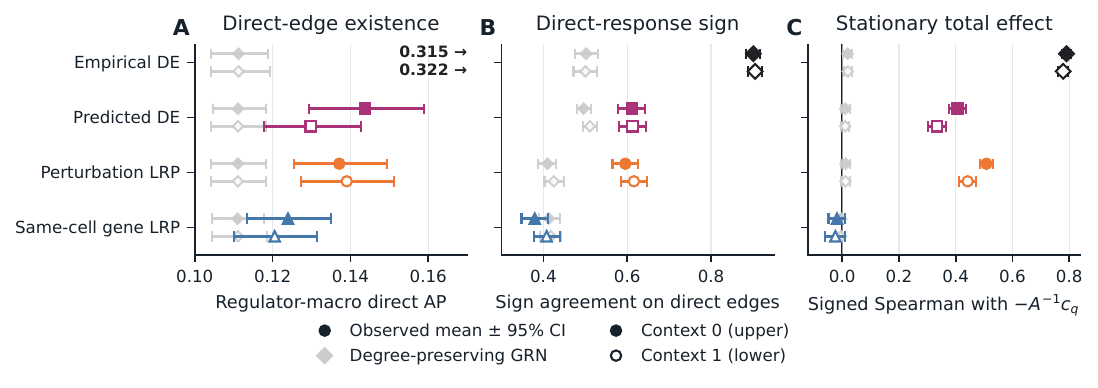}
    \caption{
    Evaluation of regulatory alignment in both in-distribution contexts (0 and 1).
    \textbf{(A)}~Binary recovery of direct edge existence, measured as average precision (AP).
    Empirical DE is clipped for better visualisation.
    \textbf{(B)}~Sign agreement with nonzero ground-truth direct responses.
    \textbf{(C)}~Spearman correlation with stationary total effects.
    Error bars for the true measurements reflect 95\% bootstrap CIs over the 128 perturbations.
    The grey control error bars are the central 95\% range of the perturbation-average evaluated against the randomised-GRN controls.
    }
    \label{fig:regulator_recovery}
    
\end{figure}

We evaluate each signature along three axes. For each perturbed gene $i_q$, target gene binary recovery is evaluated with AP (ranking genes $j \neq i_q$ with $\*s_{m, q}$ and treating non-zero entries of $\*d_q$ as positives), similarly to \cref{sec:marker_recovery} and also adopted by \citet{pratapa2020benchmarking} for GRN recovery. Sign agreement evaluates accuracy between the signs of $\*s_{m, q}$ and $\*d_q$. Alignment of a candidate signature $\*s_{m, q}$ with the total effect $\*t_q$ is measured using Spearman correlation. Metrics are computed per perturbation, then macro-averaged, and are reported in \cref{fig:regulator_recovery}.

All exploratory development used context 0 test cells, with context 1 evaluated only after the analysis protocol was fixed. Comparable results across contexts and all three metrics indicate that the findings are robust to this controlled mechanism shift rather than specific to the development context.
Overall, empirical DE shows substantially stronger regulatory alignment than the model-derived signatures, consistent with the model's moderate perturbation-prediction performance. Perturbation-LRP performs similarly to predicted DE across all three metrics, with slightly stronger alignment to total stationary effects. Gene-LRP shows modest but significant direct-edge recovery compared with the randomized-GRN control \((p=0.001)\), but does not recover response signs or total effects. Complementary analyses with extended baselines can be found in~\cref{app:extended_regulatory_alignment}.

\subsection{Attribution-based model diagnostics}
\label{sec:diagnostics}

\paragraph{Relevance concentration and prediction quality.}
We next ask whether the share of model relevance assigned to different input sources is associated with perturbation-prediction quality. For each perturbation, we compute the fraction of total absolute relevance assigned to each input source, averaged across cells and STATE batches, and correlate these fractions with Pearson-\(\Delta\) using Spearman correlation.
In-distribution, the relevance fraction assigned to the explicit perturbation input is strongly associated with prediction quality, reaching \(\rho=0.897\).
Because this association could partly reflect properties of the black-box model output, we additionally control for predicted-DE dispersion, sparsity, and magnitude. The association remains substantial after adjustment, with \(\rho=0.666\), indicating that relevance assigned to the perturbation input provides information about prediction quality beyond these output statistics.

\begin{figure}
    \centering
    \includegraphics[width=\linewidth, trim={0 10pt 0 8pt}, clip]{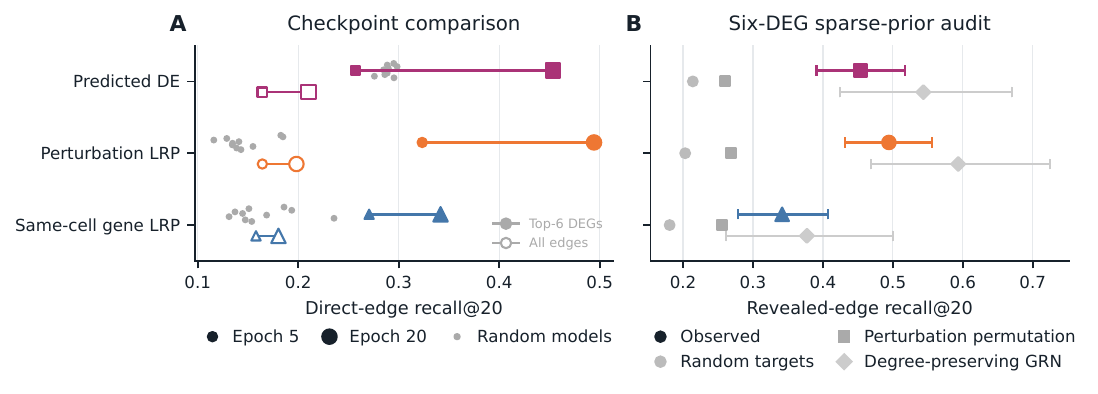}
    \caption{
    Regulator recovery under sparse prior knowledge for top-6 differentially expressed genes (DEGs), measured by recall@20.
    \textbf{(A)}~Comparing the final model checkpoint (epoch 20) vs an early checkpoint (epoch 5) and random untrained model baselines.
    Recall increases with training, and at epoch~5 LRP signatures exceed the random untrained models while predicted DE does not.
    Results under full knowledge (all direct edges) are also shown for reference as hollow markers, though the numeric scales are not directly comparable as the denominators differ.
    \textbf{(B)}~Recovering known regulatory edges, alongside randomised controls (details in \cref{sec:diagnostics}).
    }
    \label{fig:sparse_prior}
\end{figure}

\paragraph{Learning progression.}
Regulatory alignment, as measured in \cref{sec:grn_alignment}, would be useful to evaluate while training models, to track how alignment with known regulatory relationships evolves during training.
However, it is unrealistic to assume \emph{full} regulatory knowledge in practical applications.
Therefore, we simulate a scenario wherein an analyst would ascertain regulatory interactions for only the most differentially expressed genes (DEGs).
For each perturbation, we select the top-six genes (excluding the perturbed gene) with the highest training-dataset empirical DE.
The only revealed knowledge is which of the selected DEGs are directly regulated by the perturbed gene, with no interaction sign or magnitude.
Because this is a positive-vs-unknown recovery problem and AP requires labelled negatives, we adopt recall@20 as the primary metric.

\Cref{fig:sparse_prior}A shows a consistent pattern across the sparse- and full-knowledge settings: recall increases with training for all signatures and is higher for perturbation-LRP than for gene-LRP.
Further, perturbation-LRP displays a learned alignment signal by epoch 5, at which point predicted DE alignment was still not distinguishable from the random-model controls. This suggests that perturbation-specific information emerged in the perturbation pathway sooner than it was expressed in model outputs.

\paragraph{Limits of sparse auditing.}
Though we have seen that attribution signals can serve practical diagnostic uses even with no or limited ground-truth knowledge, care must be taken not to over-interpret the results as true regulatory discovery.
To illustrate this, we consider two typically well-motivated randomised controls (100 replicates each): random targets and perturbation permutation, respectively measuring the effect of DEG-panel selection and perturbation-specificity.
Indeed, sparse regulator recovery results clearly surpass these controls for all evaluated signatures in \cref{fig:sparse_prior}B.

On the other hand, none of the signatures align with the sparse ground-truth significantly better than with the degree-preserving GRNs from \cref{sec:grn_alignment}.
This implies that the signatures broadly prioritised the DEGs but did not distinguish which were direct targets of the perturbed gene,
underscoring the need for careful experimental design and controls that preserve relevant graph structure.

\section{Conclusion}
Across cell-type classification and perturbation prediction, attribution analyses provided information complementary to predictive metrics. They revealed which marker genes, contextual interactions, input pathways, and potential shortcuts supported predictions, while also exposing learning progression and prediction reliability beyond output statistics. However, biological conclusions depended on appropriate references, informative baselines, and controls matched to the data-generating process. Together, these findings establish attribution-based explainability as a powerful post-hoc tool for debugging scFMs and guiding model development by revealing the biological signals, shortcuts, and computational pathways behind their successes and failures.

Although our study shows a range of questions attribution can address, it covers selected pretrained models, linear-probe tasks, and a single perturbation model. More fundamentally, scFM interpretability is constrained by scarce biological ground-truth. Marker panels provide only partial references, while comprehensive GRNs are generally unavailable for real perturbation data, motivating our synthetic setting with fully known regulatory structure despite its limited biological complexity. 
Future work should test transfer across models, tasks, and datasets using complementary biological evidence, controlled interventions, and task-specific nulls. Broader validation could extend attribution beyond post-hoc analysis to training and model selection, supporting more robust and biologically grounded scFMs.

\clearpage
\subsection*{AI use statement}

In this work, we used generative AI tools for polishing the writing, identifying related work, and supporting research execution. 
We have not used generative AI tools to draft substantive sections of the paper, generate synthetic datasets, or make scientific conclusions.
We have reviewed all AI-assisted work. In particular, author review verified that LLM-assisted writing accurately reflected our ideas, while literature retrieval was independently validated through manual literature review.
We take responsibility for the final content of this work, including text, claims, or artifacts produced with the aid of generative AI.

\subsection*{Reproducibility statement}

For reproducibility, the CELLxGENE subsets used in this study, together with the external databases used to obtain marker-gene annotations, are described in detail in \cref{app:blood_dataset_details} and \cref{app:brain_dataset_details}. The synthetic dataset generated for the perturbation-prediction experiments is described in \cref{app:synthetic_data_details}. For the cell-type classification experiments, the complete probe-training procedure is provided in \cref{app:probe-details}. The architecture and training procedure used for the perturbation-prediction model are detailed in \cref{app:scgpt_training_details}. The methodology and implementation details of our attribution-based analyses are provided in \cref{app:lrp} and \cref{app:symbolic_xai_details}. Additional information required to reproduce the experimental results is provided in the main paper, including the experimental setup in \cref{sec:cell_typing} and \cref{sec:perturbation_prediction}.

\subsubsection*{Acknowledgments}
We would like to thank Paidamoyo Chapfuwa, Madeline Hughes, James Hall, and Jabran Zahid for sharing their domain expertise to support this research.
We also gratefully acknowledge Julius Hense and Daniel Shao for the valuable methodological discussions and feedback.

\bibliographystyle{iclr2027_conference_preprints}
\bibliography{references}

\clearpage
\appendix
\crefalias{section}{appendix}
\crefalias{subsection}{appendix}
\crefalias{subsubsection}{appendix}
\numberwithin{figure}{section}
\numberwithin{table}{section}

\section{Dataset Description}
\label{app:datasets}

\subsection{Blood PBMC Immune-Cell Benchmark}
\label{app:blood_dataset_details}
The blood benchmark contains 61{,}350 cells and 24{,}071 genes from 17 source datasets and 106 donors. Each cell has two labels, a coarse immune cell type and a fine-grained cell subtype. The final hierarchy contains 4 cell types and 22 subtypes.

\paragraph{Extraction and pre-processing.}
We constructed the blood benchmark from CZ CELLxGENE Census \citep{czi2025cellxgene}, release \texttt{2025-11-08}. The goal was to obtain a clean PBMC-like immune-cell dataset in which models could be evaluated at both cell type and subtype resolution. To reduce technical and biological heterogeneity, we restricted the source pool to human primary cells from normal blood generated with the same 10x Genomics 3$'$ v3 assay. We also required each candidate cell to have at least 500 detected genes in the Census metadata.

From this source pool, we retained only cells with informative immune-cell annotations. Broad labels that did not specify a subtype were removed, and the remaining labels were mapped into a two-level hierarchy: 4 major immune cell types and 22 nested subtypes. We then applied standard cell-level quality control to raw counts, removing cells with high mitochondrial fraction, predicted doublet status, or unusually low or high numbers of detected genes. 
Finally, we sub-sampled the filtered cells to make the subtype distribution as balanced as possible, targeting approximately 3{,}000 cells per subtype. The extracted cell types and subtypes can be seen in Table~\ref{tab:blood-celltype-hierarchy}.

\begin{table}[h]
\centering
\small
\caption{Hierarchy of cell types and cell subtypes in the dataset.}
\label{tab:blood-celltype-hierarchy}

\begin{tabularx}{0.9\textwidth}{l X}
\toprule
\textbf{Cell Type} & \textbf{Cell Subtype} \\
\midrule

T cell      & Naive CD4$^+$ T,
            Central memory CD4$^+$ T,
            Effector memory CD4$^+$ T,
            CD4$^+$ helper T,
            Cytotoxic CD4$^+$ T,
            Naive CD8$^+$ T,
            Central memory CD8$^+$ T,
            Effector memory CD8$^+$ T,
            Effector CD8$^+$ T,
            MAIT,
            $\gamma\delta$ T,
            Memory Treg,
            Naive Treg \\
\midrule

B cell      & Naive B,
            Memory B,
            Transitional B,
            Plasma cell \\
\midrule

NK cell     & CD56$^{\text{dim}}$ NK,
            CD56$^{\text{bright}}$ NK \\
\midrule

Monocyte    & Classical,
            Intermediate,
            Non-classical \\
\bottomrule
\end{tabularx}
\end{table}

\paragraph{Marker-gene extraction.}
For the blood benchmark, marker genes were curated independently of model predictions and used only as biological reference panels for downstream interpretability analyses. Candidate markers were collected from five external resources: AIFI~\citep{allen_immuneatlas, gong2025immune}, Azimuth PBMC markers~\citep{hao2021azimuth}, PanglaoDB~\citep{franzen2019panglaodb}, CellMarker~2.0~\citep{hu2023cellmarker2}, and the CELLxGENE Cell Guide~\citep{czi_cellguide}.  The candidate list was then evaluated directly on the final benchmark rather than accepted purely from the external databases.

Each candidate gene was assessed using expression patterns in the curated blood dataset, including linear counts-per-10k expression, the fraction of cells with non-zero expression, tau specificity, and one-vs-rest Wilcoxon differential-expression support.  This produced two marker panels: a cell-type-level panel for the four coarse immune cell types and a subtype-level panel for the 22 fine-grained immune populations.

\paragraph{Shared cell subtype marker genes.}
\Cref{tab:shared-marker-pairs} lists the four pairs of T cell subtypes and their shared markers, as used in the context-sensitivity analysis of \cref{sec:contextual}.

\begin{table}[h]
\centering
\caption{Blood cell subtype pairs and shared marker genes used in the context analysis.}
\label{tab:shared-marker-pairs}
\small
\setlength{\tabcolsep}{10pt}
\begin{tabular}{lll}
\toprule
\multicolumn{2}{c}{T cell subtypes} & Shared markers \\
\midrule
CM CD4 & CM CD8
    & \textit{CCR7}, \textit{IL7R} \\
EM CD4 & EM CD8
    & \textit{CCL5}, \textit{GZMA} \\
CM CD4 & EM CD4
    & \textit{CD4}, \textit{IL7R} \\
CM CD8 & EM CD8
    & \textit{CD8A}, \textit{CD8B} \\
\bottomrule
\end{tabular}
\end{table}

\subsection{Brain Broad-Class Benchmark}
\label{app:brain_dataset_details}
This benchmark contains 61{,}250 nuclei and 38{,}019 genes from 102 source datasets and 623 donors. It is balanced across 7 broad classes, with 8{,}750 nuclei per class.

\paragraph{Extraction and pre-processing.}
We constructed the brain benchmark from the same CZ CELLxGENE Census~\citep{czi2025cellxgene} release, \texttt{2025-11-08}. The goal was to obtain a clean adult human brain single-nucleus dataset for broad cell-class evaluation. To reduce technical and biological heterogeneity, we restricted the source pool to human primary brain nuclei from exact-normal samples generated with the 10x Genomics 3$'$ v3 assay.
Because the benchmark is intended to represent adult/postnatal brain, prenatal or fetal samples were excluded before class mapping.

We then mapped the remaining nuclei into seven broad classes: excitatory neurons, inhibitory neurons, astrocytes, oligodendrocytes, oligodendrocyte precursor cells, microglia/macrophages, and vascular/perivascular cells. Standard cell-level quality control was applied to raw counts, removing nuclei with insufficient detected genes, unusually high detected-gene counts within their class, or high mitochondrial fraction. The final dataset was exactly balanced across classes, with 8{,}750 nuclei per class and 61{,}250 nuclei in total.
Because the available CELLxGENE brain data are not uniformly sampled across regions, the benchmark should be interpreted as cortex-weighted rather than as a whole-brain atlas.

\paragraph{Marker-gene extraction.}
For the brain benchmark, marker genes were curated independently of model predictions and used as biological reference panels for interpretability analyses. Candidate markers were collected from Allen Brain Atlas canonical markers~\citep{allen_brainatlas, siletti2023humanbrain}, CELLxGENE Cell Guide markers~\citep{czi_cellguide}, and literature-derived canonical brain markers.

Each candidate gene present in the final dataset was evaluated using its class-level expression profile, including mean linear counts-per-10k expression, detection fraction, tau specificity, one-vs-rest Wilcoxon differential-expression rank, source provenance, primary brain-class assignment, and trust tier. Primary-class assignment was based on a robust Wilcoxon-rank rule rather than raw mean expression alone. 
The final marker panel contains 55 genes across the seven broad brain classes.

\subsection{Synthetic Dataset}
\label{app:synthetic_data_details}

For the perturbation prediction experiment, we generate a synthetic dataset based on \citet{qi2026mechanisms,lorch2026latent}. Each cell in this dataset is represented by the expression of $G=128$ genes. The data is generated from a stochastic linear dynamical system whose interactions are defined by an explicit gene regulatory network (GRN). For a cellular context \(c\), the latent expression state \(\mathbf{x}\in\mathbb{R}^G\) evolves as
\begin{equation}
\diff\mathbf{x}(t) = (\mathbf{A}_c\mathbf{x}(t) + \mathbf{b}_c + \PertVec_q)\diff t+\sigma \diff\mathbf{w}(t),
\end{equation}
where \(\mathbf{A}_c\) denotes the context-specific GRN, \(\mathbf{b}_c\) the context-specific baseline, and $\mathbf{w}(t)$ is a standard Wiener process.
Perturbation $q$ is encoded as a sparse vector $\PertVec_q$, with non-zero entries reflecting the sign and strength of the intervention on the target genes.
Also following \citet{qi2026mechanisms}, the observed expression value for each gene $g$ is sampled from a negative binomial likelihood, with mean $\operatorname{softplus}(x_g)$ and dispersion estimated from the \citet{norman19} dataset.

We generate two cellular contexts over the same 128 genes. The first context $0$ is defined by a sparse signed regulatory network, where each gene selects 10 regulators uniformly (without replacement, excluding self), and assigns each edge a random sign and magnitude uniformly between 1 and 3.  The second is obtained from the first one by rewiring 50\% of the regulatory edges in each row and modifying the context-specific baseline. The two contexts therefore share part of their regulatory structure while differing in the causal mechanisms governing gene expression. For control cells, \(\PertVec_q=\mathbf{0}\).
For perturbed cells, the intervention directly targets one of the 128 genes and its effect propagates through the corresponding GRN.
The raw expression values are subsequently normalized to a library size of \(10{,}000\) and log-transformed.

We generate 8M unperturbed control cells for pretraining and, for fine-tuning, an additional 5M controls and 1M perturbed cells across both contexts. The perturbed set contains 8,192 independently simulated cells for each of 128 randomly signed single-gene perturbations, covering all genes. Controls and perturbed cells are sampled independently from their respective stationary populations and thus have no one-to-one correspondence.
Perturbations are instantiated with random sign and random magnitude in $[2\log G, 6\log G] = [9.7, 29.1]$, bounded away from zero to prevent undetectable effects. Both datasets are divided into 80\%, 10\%, and 10\% training, validation, and test splits, respectively, stratified by context and perturbation.

\section{Cell-type classification probes}
\label{app:probe-details}

Single-cell foundation models are designed to encode cellular information in their representations, but it remains unclear where cell identity becomes linearly accessible across the network and how this depends on gene-level representation aggregation and cellular context. Understanding this is important both for characterizing the learned representations and for selecting suitable prediction targets for our attribution analyses.

\paragraph{Training details.}
We train linear probes on frozen representations extracted at approximately 25\%, 50\%, 75\%, and 100\% of the network and evaluate them on the same held-out cells. This tests whether cell identity emerges progressively or is already linearly accessible in early layers. For C2S, we construct cell-level representations using either mean pooling over non-padding tokens or the final-token representation, and for scGPT we analogously compare mean pooling with the \texttt{[CLS]} token representation. Each blood probe is a single linear layer predicting either four cell types or 22 cell subtypes, while brain probes predict one of seven cell types.

Probes are trained with AdamW and class-balanced cross-entropy. We use a separate validation split for early stopping and model selection, evaluating the test split only after selecting the final probe. The learning rate is linearly warmed up over the first 10\% of training and then follows cosine annealing. To reduce sensitivity to optimization choices, we perform a systematic grid search for each model, representation, and prediction task, evaluating every configuration across three random seeds and selecting the final probe based on validation macro-F1. The search spaces are reported in \cref{tab_probe_shared,tab_probe_models}.

\begin{table}[tbh]
\centering
\caption{Shared training and hyper-parameter search settings for linear probes.}
\label{tab_probe_shared}
\begin{tabular}{ll}
\toprule
Setting & Value \\
\midrule
Optimizer & AdamW \\
Objective & Class-balanced cross-entropy \\
Selection metric & Macro-F1 \\
Learning-rate schedule & 10\% linear warm-up, then cosine annealing \\
Weight decay & $10^{-5}, 10^{-4}, 10^{-3}$ \\
Dropout & $0.0, 0.1$ \\
Standardization & None \\
Random seeds & 0, 1, 2 \\
\bottomrule
\end{tabular}
\end{table}

\begin{table}[tbh]
\centering
\caption{Model-specific probe hyper-parameter search configurations.}
\label{tab_probe_models}
\begin{tabular}{lccccc}
\toprule
Model & Layers & Learning rate & Batch size & Epochs & Patience \\
\midrule
Pythia 410M
& 6, 12, 18, 24
& $3{\times}10^{-4}, 10^{-3}, 3{\times}10^{-3}$
& 64 & 200 & 15 \\

Pythia 1B
& 4, 8, 12, 16
& $10^{-4}, 3{\times}10^{-4}, 10^{-3}$
& 128 & 100 & 10 \\

Gemma2 2B
& 6, 13, 20, 26
& $10^{-4}, 3{\times}10^{-4}, 10^{-3}$
& 256 & 60 & 10 \\

scGPT
& 3, 6, 9, 12
& $10^{-4}, 3{\times}10^{-4}, 10^{-3}$
& 64 & 30 & 7 \\
\bottomrule
\end{tabular}
\end{table}

\paragraph{Selecting probes for attribution.}
For attribution analyses, we use probes trained on final-layer representations and select the pooling strategy separately for each model configuration and prediction task based on validation accuracy. The selected pooling strategies are reported in \cref{tab:attribution-probe-pooling}. LRP is then applied to the corresponding probe logits.

\begin{table}[t]
  \centering
  \caption{Pooling strategies selected for the final-layer probes used in attribution analyses. Each strategy is selected separately for each model, context length, and prediction task based on validation performance. A dash denotes an unevaluated combination. Context length \(K\) applies only to C2S.}
  \label{tab:attribution-probe-pooling}
  \begin{tabular}{lrccc}
    \toprule
    Model & $K$ & Blood type & Blood subtype & Brain type \\
    \midrule
    C2S Pythia 410M       & 200  & \texttt{last\_token}   & \texttt{last\_token}   & \texttt{last\_token}   \\
    C2S Pythia 1B         & 200  & \texttt{last\_token}   & \texttt{mean\_non\_pad} & \texttt{last\_token}   \\
    C2S Pythia 1B         & 1000 & \texttt{last\_token}   & \texttt{last\_token}   & \texttt{last\_token}   \\
    C2S Gemma2 2B         & 200  & \texttt{last\_token}   & \texttt{last\_token}   & \texttt{mean\_non\_pad} \\
    C2S Gemma2 2B         & 1000 & \texttt{mean\_non\_pad} & \texttt{last\_token}   & \texttt{mean\_non\_pad} \\
    scGPT blood           & ---  & \texttt{CLS}           & \texttt{CLS}           & --                     \\
    scGPT brain           & ---  & --                     & --                     & \texttt{CLS}           \\
    scGPT whole-human     & ---  & \texttt{CLS}           & \texttt{CLS}           & \texttt{CLS}           \\
    \bottomrule
  \end{tabular}
\end{table}

\paragraph{Results.}

\begin{figure}[tbh]
    \centering
    \includegraphics[width=\linewidth]{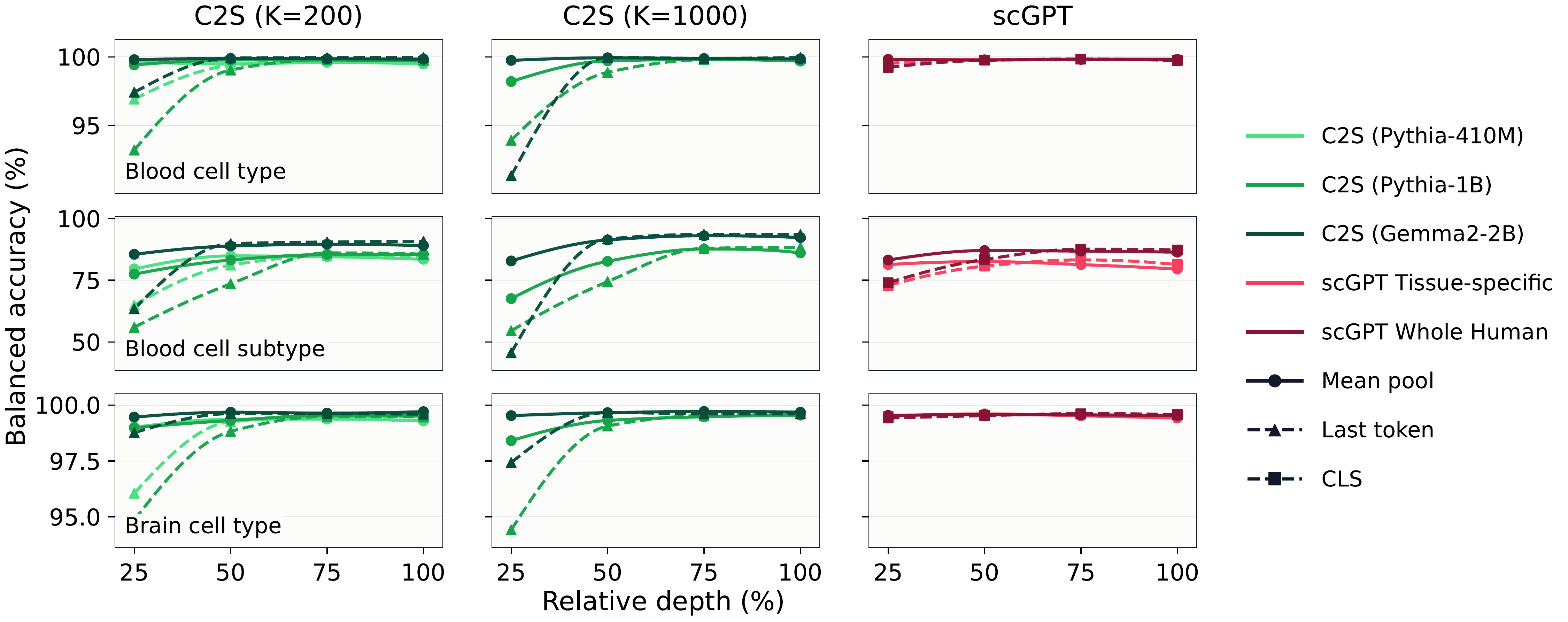}
    \caption{Probe accuracies across model depths and pooling strategies. Broad cell types are predicted more accurately than fine-grained cell subtypes. Single-token representations become increasingly predictive toward the final layers, while longer C2S input contexts generally improve accuracy. Here, \(K\) denotes the number of input genes, and \texttt{[CLS]} denotes the classification token in scGPT.}
    \label{fig:biological_probe_accuracies}
\end{figure}

\Cref{fig:biological_probe_accuracies} shows that, as expected, fine-grained cell-subtype classification is consistently more challenging than coarse-grained cell-type classification across models and layers. Cell identity also appears initially distributed across token positions, with mean-pooled representations generally providing more accessible information than single-token representations in early and intermediate layers. Toward the final layers, however, cell identity becomes increasingly concentrated in a designated summary token, namely the last token in autoregressive C2S and the \texttt{[CLS]} token in scGPT.
Finally, increasing the cellular context from \(K=200\) to \(K=1000\) genes also generally improves probe accuracy, suggesting that broader cellular context yields richer representations of cell identity.

\section{Layer-wise Relevance Propagation (LRP) Details}
\label{app:lrp}

Layer-wise Relevance Propagation \citep[LRP;][]{bach-plos15} is a framework for interpreting neural network predictions by attributing the model's output, or a quantity derived from it, to the input features. LRP assigns a relevance score $\mathcal{R}$ to each neuron that reflects its contribution to the quantity of interest and propagates these scores backward through the network using layer-specific rules. The propagation typically follows the principle of relevance conservation \citep{bach-plos15, montavon2019layerwise}, which requires the total relevance to be preserved as it is redistributed from one layer to the preceding layer. This process ultimately yields an attribution of the quantity of interest to the input features.

LRP is model-specific, as the relevance propagation rules must be adapted to the operations in specific layers. In this study, we use LRP to interpret transformer-based architectures and therefore employ LRP rules that are applicable to transformer models \citep{pmlr-v235-achtibat24a, DBLP:conf/icml/AliSEMMW22}. To determine whether LRP provides faithful explanations for our models, we compare it with Integrated Gradients, Gradient $\times$ Input, attention rollout, and a random baseline. We evaluate the faithfulness of these methods using an input perturbation approach that ranks input features according to their attribution scores \citep{Samek2015EvaluatingTV}. The ranked features are then iteratively perturbed according to either a \textit{Most Relevant First} (MoRF) or \textit{Least Relevant First} (LeRF) strategy, while tracking the output logit $f_c$ of the predicted class $c$. We denote the areas under the resulting perturbation curves by $A^F_\mathrm{MoRF}$ and $A^F_\mathrm{LeRF}$, respectively.

A faithful explanation should yield a low $A^F_\mathrm{MoRF}$, as perturbing the most relevant features is expected to strongly affect the model output. In contrast, $A^F_\mathrm{LeRF}$ should be high, as perturbing less relevant features should have a smaller effect. Following \citet{blucher2024decoupling, pmlr-v235-achtibat24a, jafari2024mambalrp}, we combine these two measures to reduce the influence of out-of-distribution perturbations and define
$
\Delta A^F = A^F_\mathrm{LeRF} - A^F_\mathrm{MoRF}.
$
A higher $\Delta A^F$ indicates a larger separation between the effects of relevant and less relevant features and therefore a more faithful explanation. We use this metric to compare the different explanation methods and select the method that most faithfully captures the behavior of the models in our case studies. We then use the same evaluation to select the most faithful combination of LRP propagation rules for each architecture. 

\paragraph{Cell typing LRP.}
The results of this analysis are reported in Table~\ref{tab:all_model_flipping_delta_mean}. LRP achieves the highest faithfulness across all evaluated architectures, outperforming the other explanation methods. We therefore use LRP as our primary attribution method. The propagation rules used for each model are reported in Table~\ref{tab:lrp_implementation_details}.

\begin{table}[t]
\centering
\caption{
Evaluating explanation methods.
Larger values indicate greater attribution faithfulness.
}
\label{tab:all_model_flipping_delta_mean}
\begin{tabular}{lccc}
\toprule
Method
& C2S (Pythia-410M)
& C2S (Gemma2-2B)
& scGPT Blood \\
\midrule
LRP (AH-rule)
& 1.198
& \textbf{1.085}
& 4.948 \\

LRP (AttnLRP-rule)
& \textbf{1.283}
& 0.945
& \textbf{5.140} \\

Gradient $\times$ Input
& -0.600
& -0.121
& 0.268 \\

Integrated Gradients
& 1.160
& 0.068
& -1.576 \\

Attention Rollout
& 0.017
& -0.002
& 0.885 \\

Random
& 0.013
& 0.005
& -0.008 \\
\bottomrule
\end{tabular}
\end{table}

\begin{table}[tbh]
\centering
\newcommand{\cmark}{\textcolor{ForestGreen}{\checkmark}}
\newcommand{\xmark}{\textcolor{red}{\texttimes}}
\caption{Propagation rules used in our LRP implementation for the three single-cell architectures considered in this study. A green tick (\cmark) indicates that the rule was applied, while a red cross (\xmark) indicates that it was not used.}
\label{tab:lrp_implementation_details}
\resizebox{\linewidth}{!}{%
\begin{tabular}{lccc}
\toprule
\textbf{Propagation Rule} 
& \textbf{C2S (Gemma2)} 
& \textbf{C2S (Pythia)} 
& \textbf{scGPT} \\
\midrule
LN-rule \citep{DBLP:conf/icml/AliSEMMW22} 
& \cmark & \cmark & \cmark \\
Identity-rule \citep{jafari2024mambalrp} 
& \cmark & \cmark & \xmark \\
$\epsilon$-rule \citep{montavon2019layerwise} 
& \cmark & \cmark & \xmark \\
Half-rule \citep{jafari2024mambalrp, Arras2019} 
& \cmark & \xmark & \xmark \\
AH-rule \citep{DBLP:conf/icml/AliSEMMW22} 
& \cmark & \xmark & \xmark \\
AttnLRP \citep{pmlr-v235-achtibat24a} 
& \xmark & \cmark & \cmark \\
\bottomrule
\end{tabular}
}
\end{table}

\paragraph{Perturbation prediction LRP.}
For the joint scGPT+STATE perturbation model, we used the LN-, identity-, and AttnLRP rules. Faithfulness was evaluated by masking genes ranked using relevance from independent cell sets.
Most-relevant-first masking produced larger prediction-shift AUCs than least-relevant-first masking ($\Delta A=0.491$ for same-cell and $0.508$ for all-cell relevance).
We additionally ablated the model's perturbation channel: condition-level prediction Pearson-$\Delta$ dropped from $0.506$ to $0.035$, and perturbation-source LRP agreed with the resulting output changes (absolute Spearman $\rho=0.814$).
Replacing AttnLRP with the AH-rule in STATE produced similar rankings ($\rho=0.999$ for same-cell and $0.946$ for all-cell relevance).

\section{Symbolic XAI details}
\label{app:symbolic_xai_details}

Symbolic XAI~\citep{schnake2025symbolic} extends conventional feature-wise attribution to explanations defined over interactions and logical relationships between input features. Let $\mathcal{N}$ denote the set of input features. The framework first represents the model prediction using a multi-order decomposition

\begin{equation}
    f(X)=\sum_{L\subseteq\mathcal{N}}\mu_L,
\end{equation}

where $\mu_L$ denotes the contribution associated with the subset of features $L$. The decomposition therefore contains both first-order terms, such as $\mu_{\{i\}}$, and higher-order terms, such as $\mu_{\{i,j\}}$, which capture contributions involving multiple input features.

In the propagation-based formulation used in this work, the multi-order terms are obtained through higher-order relevance propagation. Rather than assigning relevance only to individual input features, relevance is propagated along ordered sequences of interacting features, referred to as walks. If $R_W$ denotes the relevance associated with a walk $W$, walks involving the same set of input features are aggregated as

\begin{equation}
    \mu_L
    =
    \sum_{W:\,\mathrm{set}(W)=L} R_W.
\end{equation}

This produces relevance terms associated with individual features as well as their higher-order interactions while preserving an additive decomposition of the model prediction.

Symbolic queries are subsequently used to select and aggregate relevant terms from this decomposition. For a query $q$, the framework defines a binary function $\lambda_L(q)$ indicating whether the feature subset $L$ satisfies the logical condition expressed by $q$. The relevance associated with the query is

\begin{equation}
    A(q)
    =
    \sum_{L\subseteq\mathcal{N}}
    \eta_L\,\mu_L\,\lambda_L(q),
\end{equation}

where $\eta_L$ controls the weighting of the individual multi-order terms. A symbolic query can therefore be interpreted as a Boolean filter over the higher-order relevance decomposition, followed by aggregation of the selected terms.

In our analysis, let $S$ denote the set of marker genes shared between two
related cell subtypes, and let $O=\mathcal{N}\setminus S$ denote all other
input genes in the cellular context. We consider two symbolic queries that
differ in whether interactions between the shared marker genes and the
remaining cellular context are included.

The first query,
\begin{equation}
    q_{\mathrm{full}} = S,
\end{equation}
selects every relevance term involving at least one shared marker gene,
\begin{equation}
    \lambda_L(S)
    =
    \mathbb{1}[L\cap S\neq\emptyset].
\end{equation}
Thus, $q_{\mathrm{full}}$ includes both relevance terms formed entirely
within $S$ and higher-order terms that involve shared marker genes together
with genes from $O$. Its relevance score therefore measures the contribution
of the shared marker genes in the full cellular context.

The second query,
\begin{equation}
    q_{\mathrm{minimal}} = S \land \neg O,
\end{equation}
additionally requires that no gene outside the shared marker set participates
in the relevance term. Specifically,
\begin{equation}
    \lambda_L(S\land\neg O)
    =
    \mathbb{1}[L\cap S\neq\emptyset]\,
    \mathbb{1}[L\cap O=\emptyset].
\end{equation}
Since $O=\mathcal{N}\setminus S$, this query retains only relevance
terms composed exclusively of genes in $S$. Hence,
$q_{\mathrm{minimal}}$ provides a context-isolated measure of the relevance
of the shared marker genes, excluding their interactions with all other
genes in the cell.

Comparing $q_{\mathrm{full}}$ and $q_{\mathrm{minimal}}$ therefore allows us
to distinguish relevance arising from interactions within the shared marker
set from relevance that depends on interactions between the shared markers
and the broader cellular context.

\section{scGPT Training Details}
\label{app:scgpt_training_details}

\subsection{Pre-training}

We pretrain the scGPT backbone from scratch on the synthetic control-cell dataset described in Appendix~\ref{app:synthetic_data_details}.

\paragraph{Architecture.}
We use the original scGPT backbone. Each input cell contains expression values for 128 genes together with a \texttt{[CLS]} token, resulting in a sequence length of 129. Gene-expression values are represented as continuous normalized-log1p values, and genes with zero expression are retained explicitly. The model consists of 12 Transformer layers with an embedding dimension of 512, 8 attention heads, a feed-forward dimension of 512, and a dropout rate of 0.1. The complete pretraining model contains approximately 20.3M trainable parameters.

\paragraph{Masked expression prediction.}
Pretraining follows the masked expression-prediction setup of scGPT~\citep{cui2024scgpt}. For each batch, we sample a masking ratio uniformly from
$
\{0.25, 0.50, 0.75\},
$
corresponding to 32, 64, or 96 masked genes, respectively. Masked genes are sampled uniformly without replacement, and the model is trained to reconstruct their continuous normalized-log1p expression values from the remaining genes.

\paragraph{Reconstruction objective.}
The standard reconstruction loss is the mean-squared error over the masked genes:
\begin{equation}
\mathcal{L}_{\mathrm{micro}}
=
\frac{1}{|\mathcal{M}|}
\sum_{i \in \mathcal{M}}
(\hat{y}_i-y_i)^2,
\end{equation}
where $\mathcal{M}$ denotes the set of masked genes, $y_i$ is the target expression value of gene $i$, and $\hat{y}_i$ is the corresponding prediction.

Since training examples are unevenly distributed across genes and expression ranges, we additionally use a balanced reconstruction loss, $\mathcal{L}_{\mathrm{balanced}}$, that first balances reconstruction errors across genes within each expression stratum and then averages the resulting stratum-level losses. The primary reconstruction objective is
\begin{equation}
\mathcal{L}_{\mathrm{primary}}
=
\alpha \mathcal{L}_{\mathrm{micro}}
+
(1-\alpha)\mathcal{L}_{\mathrm{balanced}},
\end{equation}
with $\alpha=0.75$.

We further use the GEPC objective of scGPT as an auxiliary training signal. The GEPC decoder predicts masked gene-expression values from the learned cell representation. The total pretraining objective is
\begin{equation}
\mathcal{L}_{\mathrm{pre}}
=
\mathcal{L}_{\mathrm{primary}}
+
\beta \mathcal{L}_{\mathrm{GEPC}},
\end{equation}
with $\beta=0.5$.

Model selection is based only on $\mathcal{L}_{\mathrm{primary}}$, excluding the auxiliary GEPC loss.

\paragraph{Optimization.}
We optimize the model using AdamW with a learning rate of $10^{-4}$ and weight decay of $0.01$. The batch size is 32, with gradients accumulated over four batches to obtain an effective batch size of 128. Training runs for 100K optimization steps. The learning rate is linearly warmed up during the first 5K steps and subsequently follows a cosine decay schedule, reaching 10\% of its initial value at the end of training. Gradients are clipped to a maximum norm of 1.0, and mixed-precision training is enabled.

Validation is performed every 5K optimization steps using all three masking ratios. Early stopping is enabled after 50K steps with a patience of six validation evaluations, although training completes all 100K steps. The model with the best validation performance under $\mathcal{L}_{\mathrm{primary}}$ is used to initialize fine-tuning. Test evaluation is performed only after training.

\subsection{Fine-tuning}

For perturbation prediction, we augment the pretrained scGPT backbone with the State Transition architecture of STATE~\citep{adduri2025state} and jointly train the resulting model on the synthetic control--perturbed data.

\paragraph{Architecture.}
Each control-cell expression profile is processed by the scGPT gene encoder, continuous value encoder, and 12-layer Transformer backbone to obtain a \texttt{[CLS]} cell representation. This representation is projected through a two-layer basal encoder and combined with a separately projected one-hot perturbation embedding. The resulting cell representations are processed jointly by a bidirectional, non-causal STATE set Transformer.
We omit the batch encoding, as the dataset contains homogeneous cell populations for each condition.

The STATE module has hidden dimension 64, two Transformer layers, four attention heads, and dropout of 0.1. Both the basal-state and perturbation encoders contain two layers, followed by a two-layer output decoder. We additionally use a learned context token together with an auxiliary context classifier. The decoder predicts post-perturbation expression values for the same 128 genes, with a ReLU activation applied to the output.

\paragraph{Fine-tuning objective.}
Since predicted and observed perturbed cells do not have a one-to-one correspondence, we follow STATE~\cite{adduri2025state} and optimize the model at the population level using an Energy distance between the predicted and observed perturbed cell sets. The objective additionally regularizes the predicted change relative to the control population and includes an auxiliary context-classification loss:
\begin{equation}
\mathcal{L}_{\mathrm{post}}
=
\mathcal{L}_{\mathrm{Energy}}
+
\lambda_{\Delta}\mathcal{L}_{\Delta}
+
\lambda_{\mathrm{context}}\mathcal{L}_{\mathrm{context}}.
\end{equation}

Here, $\mathcal{L}_{\mathrm{context}}$ denotes the cross-entropy loss of the auxiliary context classifier. We set

$$
\lambda_{\Delta}=10^{-5}
\quad \text{and} \quad
\lambda_{\mathrm{context}}=0.1.
$$

The Energy distance is evaluated over four randomized gene chunks with a blur parameter of $0.05$.

\paragraph{Optimization.}
Fine-tuning proceeds in two stages. During the first five epochs, the pretrained scGPT backbone is frozen and only the STATE parameters are optimized. We then unfreeze the complete scGPT gene encoder, value encoder, and Transformer backbone and jointly optimize scGPT and STATE for up to 15 additional epochs.

STATE parameters are optimized with a learning rate of $10^{-4}$, while the pretrained scGPT parameters use a smaller learning rate of $10^{-6}$. We use the Adam optimizer. Gradients are clipped to a maximum norm of 1.0. Each training example comprises a set of 64 cells, and each training step processes 2 sets. After unfreezing the scGPT backbone, early stopping with a patience of five validation evaluations is applied, and the model with the best validation performance is used for final evaluation.

\section{Perturbation prediction metrics}
\label{app:perturb_metrics}

We evaluate the predictive performance of our perturbation model using metrics from \cite{vinastorne2026systema}.
For a fixed cellular context, let $G$ denote the number of genes and let
$\mathcal{P}$ denote the set of evaluated perturbations. For perturbation
$p\in\mathcal{P}$, we define
\begin{itemize}
    \item $\boldsymbol{\mu}_p\in\mathbb{R}^{G}$: the observed mean
    post-perturbation expression profile,
    \item $\widehat{\boldsymbol{\mu}}_p\in\mathbb{R}^{G}$: the predicted mean
    post-perturbation expression profile,
    \item $\boldsymbol{\mu}_0\in\mathbb{R}^{G}$: the observed control centroid.
\end{itemize}
All metrics are calculated separately for each perturbation and context and are
then averaged across the evaluated conditions.

\paragraph{Perturbed reference.}
To reduce the influence of a generic control-to-perturbed expression shift, we
use the equal-weighted mean of the training perturbation centroids as reference:
\begin{equation}
    \mathbf{r}
    = \frac{1}{|\mathcal{P}_{\mathrm{train}}|}
      \sum_{q\in\mathcal{P}_{\mathrm{train}}}
      \boldsymbol{\mu}^{\mathrm{train}}_q.
\end{equation}
The observed and predicted perturbation-specific shifts are therefore
\begin{equation}
    \boldsymbol{\Delta}_p = \boldsymbol{\mu}_p-\mathbf{r},
    \qquad
    \widehat{\boldsymbol{\Delta}}_p
    = \widehat{\boldsymbol{\mu}}_p-\mathbf{r}.
\end{equation}

\paragraph{Perturbed-reference Pearson $\Delta$.}
Pearson $\Delta$ measures whether the predicted and observed response vectors
have the same relative pattern across genes:
\begin{equation}
    \mathrm{Pearson}\Delta_p
    =
    \frac{
        \sum_{g=1}^{G}
        (\Delta_{p,g}-\overline{\Delta}_{p,g})
        (\widehat{\Delta}_{p,g}-\overline{\widehat{\Delta}}_{p,g})
    }{
        \sqrt{\sum_{g=1}^{G}(\Delta_{p,g}-\overline{\Delta}_{p,g})^2}
        \sqrt{\sum_{g=1}^{G}
        (\widehat{\Delta}_{p,g}-\overline{\widehat{\Delta}}_{p,g})^2}
    }.
\end{equation}
It ranges from $-1$ to $1$, with larger values indicating better recovery of
the direction and relative structure of the perturbation response. Because
correlation is invariant to rescaling, it does not assess whether the predicted
magnitudes are correct.

\paragraph{Perturbed-reference Pearson $\Delta_{20}$.}
For each perturbation, let $\mathcal{D}_p$ be the 20 genes with the largest
observed absolute changes from control:
\begin{equation}
    \mathcal{D}_p
    = \operatorname{Top20}_{g}
      \left|\mu_{p,g}-\mu_{0,g}\right|.
\end{equation}
Pearson $\Delta_{20}$ is the same Pearson correlation restricted to
$g\in\mathcal{D}_p$:
\begin{equation}
    \mathrm{Pearson}\Delta_{20,p}
    = \operatorname{corr}
      \left(
        \boldsymbol{\Delta}_{p,\mathcal{D}_p},
        \widehat{\boldsymbol{\Delta}}_{p,\mathcal{D}_p}
      \right).
\end{equation}
This metric focuses on the strongest and most biologically prominent expression
changes, which can otherwise be diluted by many weakly affected genes.

\paragraph{Root mean-squared error.}
RMSE measures the absolute numerical discrepancy between predicted and observed
post-perturbation expression:
\begin{equation}
    \mathrm{RMSE}_p
    = \sqrt{
        \frac{1}{G}
        \sum_{g=1}^{G}
        \left(\widehat{\mu}_{p,g}-\mu_{p,g}\right)^2
      }.
\end{equation}
The corresponding top-20 metric is
\begin{equation}
    \mathrm{RMSE}_{20,p}
    = \sqrt{
        \frac{1}{|\mathcal{D}_p|}
        \sum_{g\in\mathcal{D}_p}
        \left(\widehat{\mu}_{p,g}-\mu_{p,g}\right)^2
      }.
\end{equation}
Lower values are better. Unlike Pearson correlation, RMSE is sensitive to the
magnitude of the prediction and is independent of the choice of expression
reference. RMSE therefore prevents a model from receiving a high score merely
for predicting the correct direction with an incorrect response magnitude.

\paragraph{Centroid accuracy.}
Centroid accuracy evaluates whether a predicted profile is closer to its
correct observed perturbation centroid than to the centroids of other
perturbations. Using Euclidean distance, it is defined as
\begin{equation}
    \mathrm{CA}_p
    = \frac{1}{|\mathcal{P}|-1}
      \sum_{q\in\mathcal{P},\,q\neq p}
      \mathbb{I}\!\left[
        \left\|\widehat{\boldsymbol{\mu}}_p-
        \boldsymbol{\mu}_p\right\|_2
        <
        \left\|\widehat{\boldsymbol{\mu}}_p-
        \boldsymbol{\mu}_q\right\|_2
      \right],
\end{equation}
where $\mathbb{I}[\cdot]$ is the indicator function. A score of $1$ means that
the correct perturbation centroid is closer than every competing centroid,
whereas a score near $0.5$ corresponds to approximately random ranking. This
metric assesses whether the model reconstructs the geometry and identity of the
perturbation landscape.

\paragraph{Systematic variation.}
Let the true control-referenced shift of perturbation $p$ be
\begin{equation}
    \mathbf{s}_p=\boldsymbol{\mu}_p-\boldsymbol{\mu}_0,
\end{equation}
and let the average perturbation shift be
\begin{equation}
    \overline{\mathbf{s}}
    = \frac{1}{|\mathcal{P}|}
      \sum_{p\in\mathcal{P}}\mathbf{s}_p.
\end{equation}
Systematic variation is the mean cosine similarity between each
perturbation-specific shift and the average shift:
\begin{equation}
    \mathrm{SV}
    = \frac{1}{|\mathcal{P}|}
      \sum_{p\in\mathcal{P}}
      \frac{
        \mathbf{s}_p^{\mathsf{T}}\overline{\mathbf{s}}
      }{
        \|\mathbf{s}_p\|_2\,\|\overline{\mathbf{s}}\|_2
      }.
\end{equation}
A high value indicates that many perturbations share one generic response
direction, which can inflate conventional correlation scores. A value near zero
indicates more heterogeneous, perturbation-specific responses. Systematic
variation is a dataset diagnostic rather than a direct model-performance metric.

\paragraph{Perturbed-mean baseline.}
We compare the model with a non-parametric baseline that predicts the same
cell-weighted mean training perturbation profile for every perturbation:
\begin{equation}
    \mathbf{b}
    = \frac{1}{N_{\mathrm{pert}}^{\mathrm{train}}}
      \sum_{i\in\mathrm{perturbed\ train\ cells}}\mathbf{x}_i,
    \qquad
    \widehat{\boldsymbol{\mu}}^{\mathrm{baseline}}_p=\mathbf{b}.
\end{equation}
The model and baseline are evaluated with identical RMSE, RMSE$_{20}$, and
centroid-accuracy calculations. Beating this baseline demonstrates that the
model captures perturbation-specific information rather than only the average
perturbed-cell state.

\section{Degree-preserving random GRN control.}
\label{app:random_controls}
To test whether regulatory alignment exceeds what could arise from network topology alone, we generate, for each context $c$, 999 randomized versions of its GRN $A_c$ by rewiring edges while preserving each gene's in- and out-degree. Candidate signatures are kept fixed and evaluated against each randomized GRN using the same recovery metrics as for the ground-truth network. Each metric is computed per perturbation and macro-averaged across perturbations.

For each signature and metric, we compute a one-sided empirical \(p\)-value as
\begin{equation}
    p =\frac{1 + \sum_{r=1}^{999} \mathbb{I}\!\left[v_r \geq {v}_{\mathrm{obs}}\right]}{1000},
\end{equation}
where \(v_r\) is the score obtained for randomized network \(r\) and \(v_{\mathrm{obs}}\) is the score for the ground-truth GRN. Thus, a small \(p\)-value indicates that the observed alignment is rarely matched or exceeded by degree-preserving randomized networks. With 999 randomizations, the minimum attainable empirical \(p\)-value is \(0.001\).

\section{Extended results}
\label{app:extended_results}

\subsection{Perturbation prediction results}

The scGPT+STATE model achieves a centroid accuracy of
$0.962$, indicating strong preservation of perturbation identity. It also reaches a Pearson $\Delta_{20}$ of $0.704$, showing good recovery of the dominant gene-response directions. Across all genes, the model obtains an RMSE of $0.855$, compared with $0.606$ for the baseline, while on the 20 most strongly affected genes it improves RMSE$_{20}$ over baseline from $1.256$ to $1.123$.
The systematic variation score of $0.147$ is low, suggesting that these results are not driven solely by a shared generic perturbation response. Overall, the
model recovers perturbation identity and dominant response patterns well, but
exact expression values are not consistently well recovered.

\subsection{Regulatory alignment}
\label{app:extended_regulatory_alignment}
In this section, we provide complementary results for the regulatory-alignment analysis using the previously introduced baselines together with additional variants. Beyond same-cell gene LRP, \textit{all-cell gene LRP} propagates relevance to input genes across all control cells in the STATE input set and aggregates their contributions.

\paragraph{Counterfactual baselines.}
As counterfactual baselines, we directly modify the model inputs and measure how much the predicted expression profile changes. We either mask a gene embedding or ablate the perturbation embedding, providing black-box measures of how strongly the model depends on each input.
Gene-level counterfactuals are reduced in a similar manner to gene-LRP signatures, by keeping only the intervention on input gene $i_q$.

\begin{figure}[h]
\centering
\includegraphics[width=\linewidth]{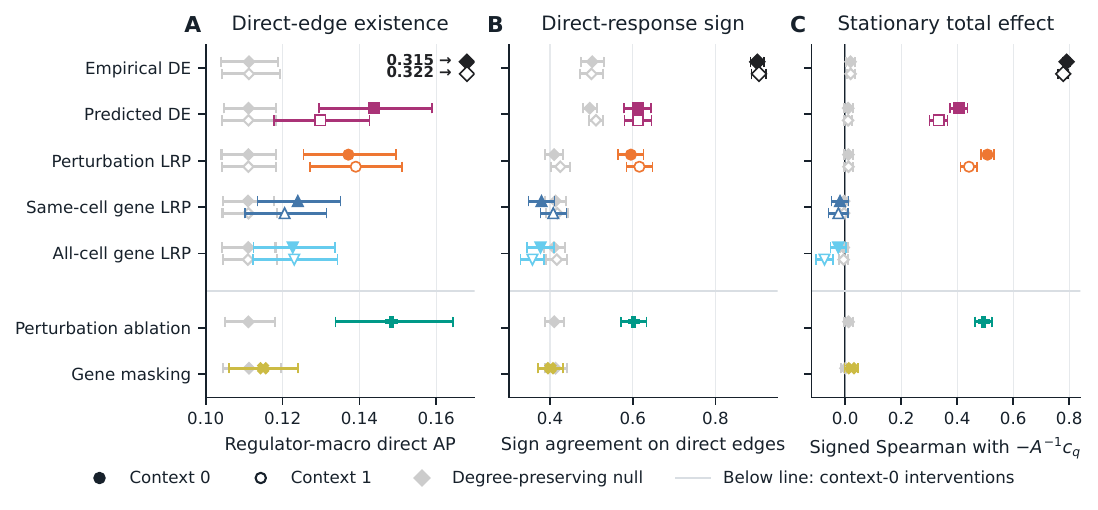}
\caption{
Evaluation of regulatory alignment in both in-distribution contexts (0 and 1).
\textbf{(A)}~Binary recovery of direct edge existence, measured as average precision (AP).
Empirical DE is clipped for better visualisation.
\textbf{(B)}~Sign agreement with nonzero ground-truth direct responses.
\textbf{(C)}~Spearman correlation with stationary total effects.
Error bars for the true measurements reflect 95\% bootstrap CIs over the 128 perturbations.
The grey control error bars are the central 95\% range of the perturbation-average evaluated against the randomised-GRN controls.
This is an extension of \cref{fig:regulator_recovery}, including all-cell gene LRP as well as interventional signatures: perturbation ablation and gene masking.
}
\label{fig:full_regulator_recovery}
\end{figure}

The extended analysis in~\cref{fig:full_regulator_recovery} largely confirms the main findings. Same-cell and all-cell gene LRP behave similarly, showing modest direct-edge recovery but little alignment with response signs or stationary total effects, indicating that aggregating relevance across all control cells provides no clear advantage. In contrast, perturbation ablation closely tracks perturbation LRP across all three metrics and achieves strong model-derived regulatory alignment, supporting the relevance assigned to the explicit perturbation input as a meaningful measure of model dependence. Gene masking remains close to the degree-preserving null, with weak direct-edge recovery and little sign or total-effect alignment. These trends are consistent across the two evaluated contexts where both are available.

\end{document}